\documentclass[11pt]{article}
\usepackage[final]{amlc_guide}
\usepackage[utf8]{inputenc}
\usepackage[T1]{fontenc}
\usepackage{hyperref}
\usepackage{url}
\usepackage{booktabs}
\usepackage{multirow}
\usepackage{amsfonts}
\usepackage{nicefrac}
\usepackage{microtype}
\usepackage{graphicx}
\usepackage{xcolor}
\usepackage{amsmath}
\usepackage{amssymb}
\usepackage{geometry}
\usepackage{fancyhdr}
\usepackage{titlesec}
\usepackage{tikz}
\usetikzlibrary{positioning,arrows.meta}
\usepackage{enumitem}
\usepackage{float}

\usepackage{listings}

\makeatletter
\renewcommand{\maketitle}{
    \begin{center}
        \rule{\textwidth}{4pt}
        \vskip 0.25in
        {\fontsize{17}{20}\selectfont\bfseries \@title \par}
        \vskip 0.25in
        \rule{\textwidth}{1pt}
        \vskip 1.5em
        {\large \@author \par}
        \vskip 1em
    \end{center}
    \vskip 2em
}
\makeatother

\titleformat{\section}{\fontsize{12}{14}\selectfont\bfseries}{\thesection}{1em}{}
\titleformat{\subsection}{\fontsize{10}{12}\selectfont\bfseries}{\thesubsection}{1em}{}
\titleformat{\subsubsection}{\fontsize{10}{12}\selectfont\bfseries}{\thesubsubsection}{1em}{}

\title{Exploring Diagnostic Prompting Approach for Multimodal LLM-based Visual Complexity Assessment: A Case Study of Amazon Search Result Pages}

\author{
  \begin{tabular*}{\textwidth}{@{\extracolsep{\fill}}ccc@{}}
    Divendar Murtadak & Yoon Kim & Trilokya Akula\\
    \small\texttt{divenm@amazon.com} & \small\texttt{yoonskim@amazon.com} & \small\texttt{atriloky@amazon.com}
  \end{tabular*}
}

\begin{document}

\maketitle

% Include sections
\begin{abstract}
This study investigates whether diagnostic prompting can improve Multimodal Large Language Model (MLLM) reliability for visual complexity assessment of Amazon Search Results Pages (SRP). We compare diagnostic prompting with standard gestalt principles-based prompting using 200 Amazon SRP pages and human expert annotations. Diagnostic prompting showed notable improvements in predicting human complexity judgments, with F1-score increasing from 0.031 to 0.297 (+858\% relative improvement), though absolute performance remains modest (Cohen's $\kappa$ = 0.071). The decision tree revealed that models prioritize visual design elements (badge clutter: 38.6\% importance) while humans emphasize content similarity, suggesting partial alignment in reasoning patterns. Failure case analysis reveals persistent challenges in MLLM visual perception, particularly for product similarity and color intensity assessment. Our findings indicate that diagnostic prompting represents a promising initial step toward human-aligned MLLM-based evaluation, though failure cases with consistent human-MLLM disagreement require continued research and refinement in prompting approaches with larger ground truth datasets for reliable practical deployment.

\end{abstract}

\section{Introduction}

In e-commerce environments like Amazon SRP, interfaces continuously evolve with hundreds of experiments running simultaneously. At Amazon's scale, with millions of SRP configurations generated daily, visual complexity assessment becomes a critical scalability challenge. Visual complexity, defined as the degree of cognitive burden imposed by the user interface, significantly impacts customer experience and decision-making.

Amazon's user research and A/B testing generate valuable behavioral data, yet the massive scale of search page variations necessitates complementary automated assessment approaches. Human annotation studies provide high-quality insights but require significant resources to scale, while business metrics from A/B testing often fail to capture nuanced perceptual aspects of visual complexity.

Multimodal Large Language Models (MLLMs) present a promising alternative for scalable visual assessment. However, MLLMs suffer from hallucination and interpretation problems in subjective assessment tasks, and full fine-tuning remains impractical due to computational costs. This motivates the exploration of diagnostic reasoning prompting, which guides models through structured analytical processes to enhance reliability and interpretability while mimicking human reasoning processes for visual complexity classification.

This study systematically evaluates diagnostic prompting effectiveness for MLLM-based visual complexity assessment. 

First, we collected human judgments on visual complexity for 200 Amazon search result pages, creating reliable ground truth data for evaluation.

Second, we tested two approaches: standard prompting (asking MLLMs to judge complexity directly) versus diagnostic prompting (guiding MLLMs through structured questions). Diagnostic prompting demonstrated notable relative improvements in F1-score from 0.031 to 0.297 (858\% relative improvement).

Third, we analyzed how MLLMs answered the diagnostic questions using decision trees to understand their reasoning. We found MLLMs focus heavily on visual design elements like badge clutter (38.6\% of their decision-making), while humans prioritize whether products look too similar. These patterns are somewhat aligned but show clear differences in how humans and MLLMs evaluate complexity.

Fourth, examining specific failure cases revealed persistent gaps. Even with diagnostic prompting, MLLMs struggle with cases involving product similarity and color intensity - areas where human evaluators unanimously disagree with MLLM assessments.

For Amazon's experimentation needs, diagnostic prompting shows meaningful improvements over standard approaches. However, reliability gaps require additional research with larger datasets before practical deployment.
\section{Customer Problem}

The concept of 'search page clutter' has gained significant traction across social media and blog posts, with customers increasingly criticizing Amazon search pages as visually overwhelming. Amazon search result pages must balance multiple interface elements including advertising placements, navigation filters, promotional badges, and organic search results while maintaining usability and relevance.

\begin{figure}[htbp]
\centering
\includegraphics[width=0.3\textwidth]{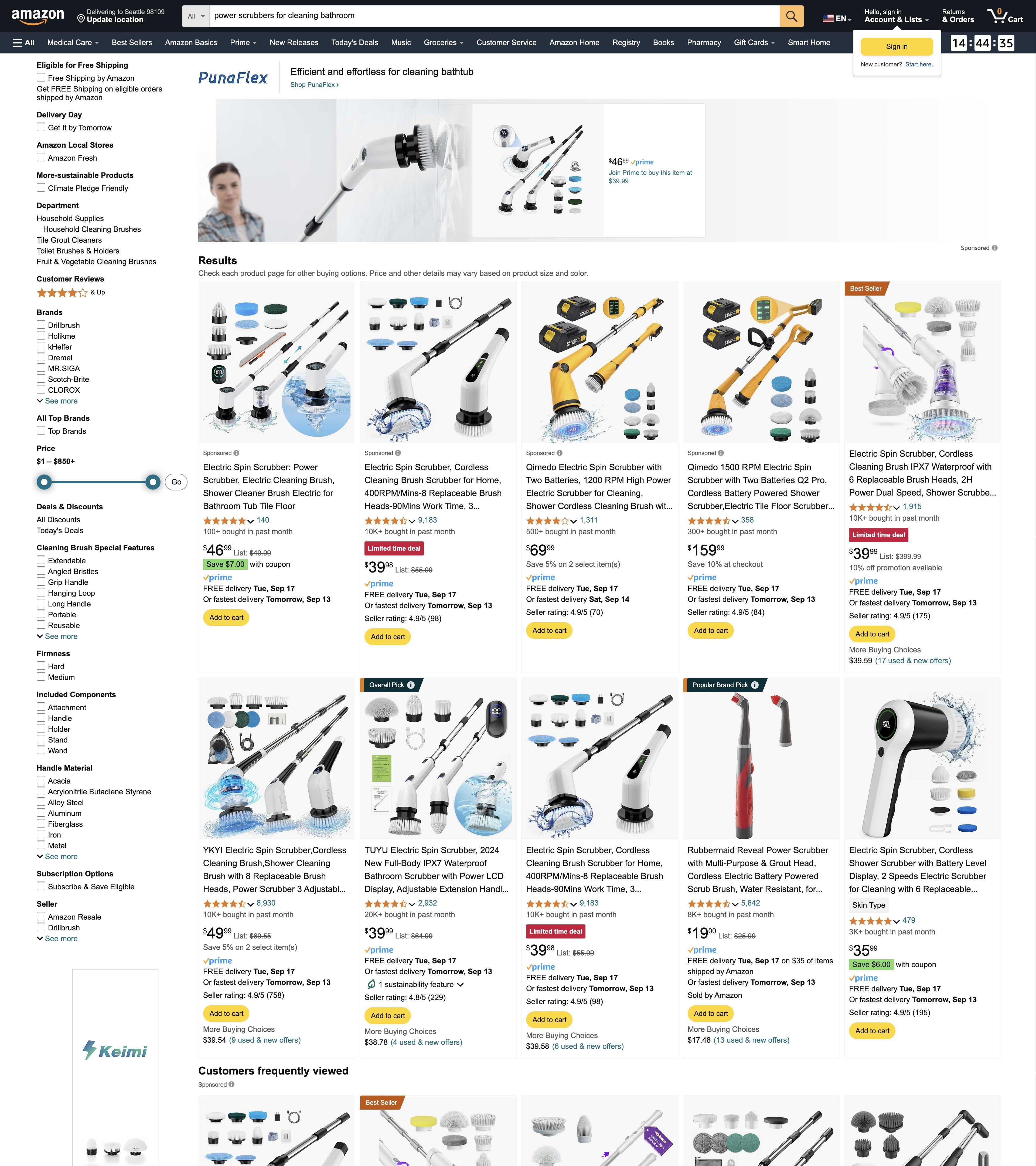}
\includegraphics[width=0.3\textwidth]{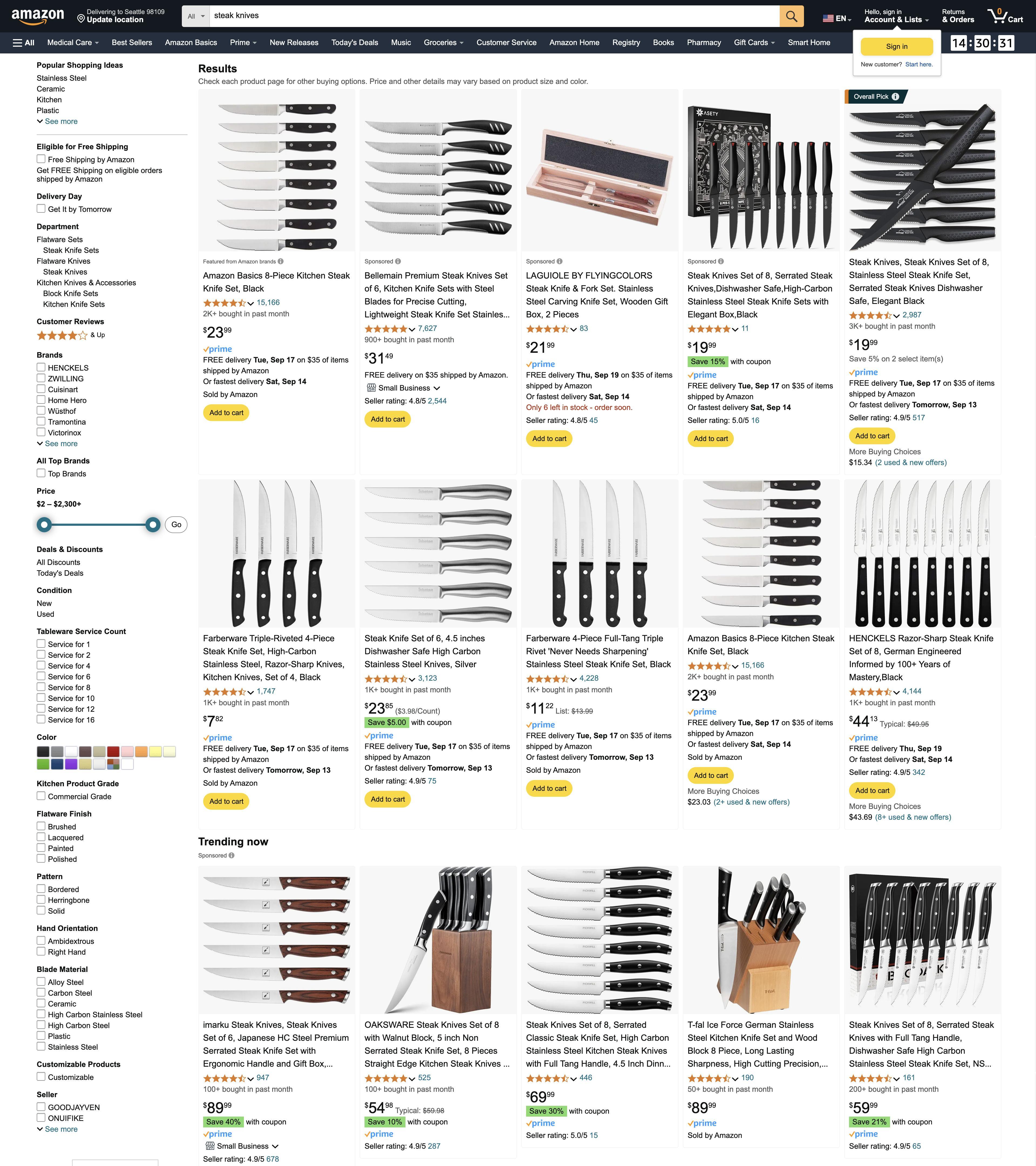}
\includegraphics[width=0.3\textwidth]{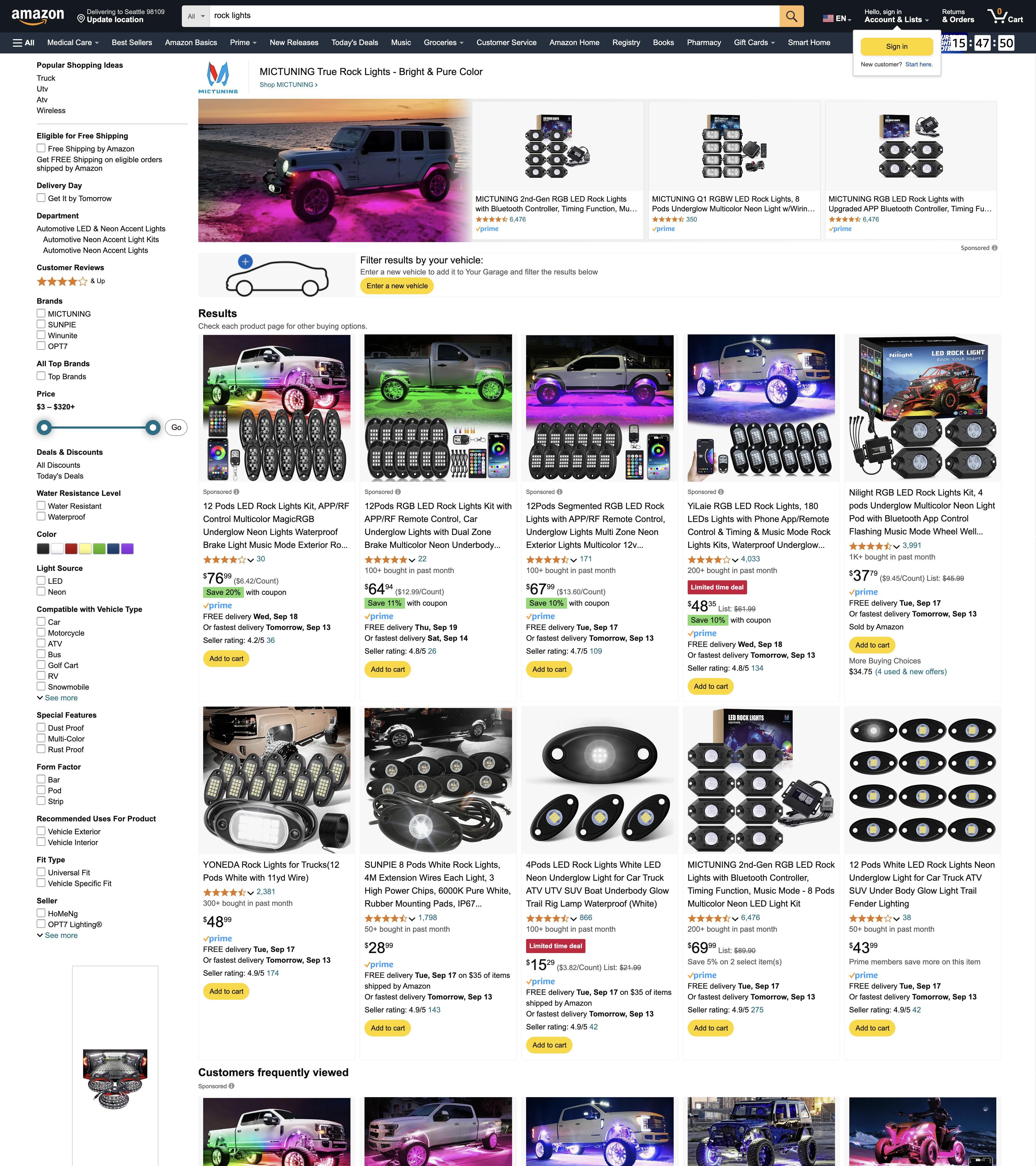}
\caption{Visual complexity challenges in Amazon search interfaces showing (left) similarity-based complexity where products appear nearly identical, (center) content overload with excessive information density, and (right) visual overload with competing visual elements.}
\label{fig:complexity_examples}
\end{figure}

As illustrated in Figure~\ref{fig:complexity_examples}, visual complexity manifests in multiple problematic forms: similarity-based confusion where products appear nearly identical, content overload with excessive information per listing, and visual chaos from overwhelming colors and clustered elements.

Without standardized assessment frameworks, interface optimization efforts may inadvertently increase complexity in some areas while addressing issues in others, highlighting the need for systematic evaluation approaches. Current evaluation methods cannot scale to assess the millions of SRP configurations generated daily across diverse product categories and personalization contexts. 

This creates a need for reliable quality metrics that capture the whole page experience, combined with scalable automated techniques for measuring visual complexity consistently across all search page variations. Such measurement approaches must align with human judgment while providing the throughput necessary to support continuous experimentation and optimization at scale. Given the subjective nature of visual complexity assessment, Multimodal Large Language Models present a promising approach for bridging human-like perceptual evaluation with the scalability requirements of Amazon's experimentation ecosystem.

\section{Related Work}

Visual complexity research has evolved from handcrafted features to deep learning architectures, with recent advances transitioning to Vision Transformers for better global visual relationships \cite{khan2023survey, scabini2024comparativesurvey}. While computational approaches explore various feature extraction methods \cite{machado2015computerized, forsythe2011predicting, donderi2006visual}, few systematically integrate perceptual psychology with modern multimodal AI frameworks \cite{ganu2025doyouseeme, werner2024pov}. Cognitive load theory demonstrates visual complexity's impact on information processing \cite{sweller1988cognitive, paas2003cognitive}, yet existing methods often fail to capture subjective, context-dependent complexity perception \cite{sarıtaş2025complexity}.

Gestalt psychology principles \cite{wertheimer1923gestalt, kohler1947gestalt, koffka1935principles} provide theoretical foundations for understanding visual complexity through organizational patterns like closure, similarity, and proximity. Recent computational work formalizes these principles using mathematical frameworks \cite{lin2024gestalt, soydaner2024finding}, though integration with modern AI evaluation systems remains underexplored \cite{markelius2025cognitivescience}.

Traditional evaluation approaches face fundamental limitations. User studies are constrained by small groups and limited duration \cite{tuch2012visual, michailidou2008visual}, visual clutter measurements fail to capture subjective perception \cite{rosenholtz2007measuring, moumoulidou2025perceptionawaresampling}, and neither scales to modern platforms \cite{fekete2022scalability, team2023gemini}. While MLLMs like GPT-4V and Claude show promise for visual evaluation \cite{xiong2024llavacritic, wang2024all, liu2023visual}, significant gaps persist in human alignment and consistency \cite{yerramilli2025huemanity, zhang2025aligning}. Their "black box" reasoning lacks systematic structure for consistent evaluation criteria \cite{ji2025interpretable, wan2025survey}.

Previous research focuses on improving MLLM accuracy through better prompting \cite{shahid2025future}, but little attention addresses systematic, interpretable reasoning for subjective visual evaluation \cite{cheng2025visually, qiang2025verb}. Our work addresses this gap by grounding MLLM evaluation in Gestalt principles through structured diagnostic questioning \cite{chaudhari2025multimodal}, providing a framework that simultaneously addresses scalability, interpretability, and reliability challenges.

\section{Data and Methodology}

\subsection{Data}
Our analysis employs 200 Amazon SRPs that represent high-traffic queries across Hardlines (117 queries, 58\%), Consumables (50 queries, 25\%), and Softlines (33 queries, 17\%). This distribution captures diverse visual complexity patterns from electronics to clothing interfaces. All samples have complete annotation coverage with no missing data, including standard prompting, diagnostic prompting, and human expert assessments captured under standardized conditions. 
Human ground truth annotations were established through consensus involving 4-7 volunteer annotators with search domain expertise per query. Annotators were provided with standardized evaluation guidelines and instructed to follow a structured rubric for consistency in complexity assessments. Final annotations represent binary complexity judgments along with complexity driver selection from predefined checklists specifically curated based on Amazon SRP page layout elements to ensure alignment with the visual complexity factors inherent in Amazon search interfaces.

\subsection{Experimental Design}
We experimented with different standard prompting approaches including zero-shot, single-shot, and few-shot learning configurations, all of which demonstrated similar performance compared to ground truth annotations; therefore, we adopted the standard single-shot approach to maintain experimental simplicity and clarity. Our experimental design compares this standard single-shot MLLM evaluation against our diagnostic reasoning approach using human expert consensus annotations as ground truth. We employed prompt-based evaluation rather than fine-tuning the MLLM on human annotations, as fine-tuning would require substantial computational resources and specialized infrastructure, making prompt engineering a more practical approach for exploring diagnostic reasoning effectiveness. The diagnostic framework was developed through an iterative process using sample human annotations to identify the most discriminative complexity factors. We transformed human reasoning patterns into 25 structured questions using Amazon SRP terminology, with detailed development methodology provided in Appendix~\ref{app:question_development}). This enables assessment through specific, observable factors rather than abstract judgments. Standard Gestalt-based and diagnostic reasoning prompts are provided in Appendix~\ref{app:standard_prompt} and Appendix~\ref{app:diagnostic_prompt}, with detailed rationale, methodology, and comparative analysis in Appendices~\ref{app:rationale_for_diagnostic_prompt}, \ref{app:question_development}, and \ref{app:comparative_analysis}.

\subsection{Implementation}
Our implementation employs Claude Sonnet 3.7 which provided the best performance across other SOTA MLLMs available at the time of experiment. The experimental design compares two prompting strategies: standard single-shot complexity assessment versus systematic evaluation using our diagnostic reasoning prompt with binary responses and evidence justification. Performance evaluation employs precision, recall, and F1-score metrics assessed against human expert consensus labels. Decision tree analysis with 5-fold stratified cross-validation provides post-hoc interpretability by analyzing MLLM diagnostic responses to understand reasoning patterns and derive feature importance rankings. All experiments were conducted with consistent temperature settings (0.1) and deterministic sampling parameters to ensure reproducibility across evaluation runs.

\section{Results}

\subsection{Statistical Validation and Performance Analysis}

Table~\ref{tab:confusion_matrix} below presents confusion matrices that reveal differences in classification patterns between standard and diagnostic prompting approaches. Standard prompting correctly classifies only 1 out of 59 truly complex images (1.7\% recall), while diagnostic prompting identifies 15 out of 59 complex cases (25.4\% recall). While diagnostic prompting increases false positives from 4 to 26, this trade-off improves MLLM's ability to detect actual complexity cases.

\begin{table}[H]
\centering
\caption{Confusion Matrices Comparing Human Annotation with MLLM Annotation}
\label{tab:confusion_matrix}
\begin{tabular}{clccc}
\toprule
\multicolumn{1}{l}{}                                                        &            & \multicolumn{3}{c}{Standard Prompting}      \\ \cline{3-5} 
\multicolumn{1}{l}{}                                                        &            & Complex & Not Complex & Total \\ \hline
\multirow{8}{*}{\begin{tabular}[c]{@{}c@{}}Human\\ Annotation\end{tabular}} & Complex &    1        &    58  &   59       \\
                                                                            & Not Complex &    4      &   137     &   141     \\
                                                                            & Total   &    5      &   195    &    200   \\
                                                                            &            &            &            &          \\ \cline{3-5} 
                                                                            &            & \multicolumn{3}{c}{Diagnostic Prompting}       \\ \cline{3-5} 
                                                                            & Complex &    15    &   44         &   59       \\
                                                                            & Not Complex &     26   &    115        &   141       \\
                                                                            & Total   &    41    &    159        &   200    \\
\bottomrule
\end{tabular}
\end{table}

Table~\ref{tab:performance_metrics} quantifies the performance differences observed between prompting approaches. Diagnostic prompting shows an F1-score increase from 0.031 to 0.297. Most notably, recall improved from 0.017 to 0.250, though starting from a very low baseline. Precision improved from 0.200 to 0.366. The Cohen's Kappa improvement from -0.016 to 0.071 represents a shift from worse-than-random to positive human-MLLM alignment. These results suggest diagnostic prompting shows promise for structured visual assessment, though substantial performance variability indicates current models require further development for reliable deployment.

\begin{table}[H]
\centering
\caption{Classification Performance Comparison: Standard vs Diagnostic Prompting}
\label{tab:performance_metrics}
\begin{tabular}{lcccc}
\toprule
\textbf{Method} & \textbf{Precision} & \textbf{Recall} & \textbf{F1-Score} & \textbf{Cohen's Kappa} \\
\midrule
Standard Prompting & 0.200 & 0.017 & 0.031 & -0.016 \\
Diagnostic Prompting & 0.366 & 0.250 & 0.297 & 0.071 \\
\midrule
\textbf{Absolute Improvement} & \textbf{+0.166} & \textbf{+0.233} & \textbf{+0.266} & \textbf{+0.087} \\
\textbf{Relative Improvement} & \textbf{+83\%} & \textbf{+1,371\%} & \textbf{+858\%} & \textbf{--} \\
\bottomrule
\end{tabular}
\end{table}

While these relative improvements are encouraging, it is important to contextualize the absolute performance levels. The F1-score of 0.297 and Cohen's $\kappa$ of 0.071 indicate that performance remains well below levels typically required for practical deployment ($\kappa$ > 0.6 for substantial agreement). These results suggest diagnostic prompting represents an initial step toward improved MLLM evaluation rather than a deployment-ready solution.

\subsection{Model Interpretability and Feature Analysis}

Decision tree models trained on diagnostic responses reveal distinct reasoning patterns between humans and MLLMs that provide interpretable complexity assessment rules. Table~\ref{tab:decision_rules} demonstrates how MLLM diagnostic prompting follows systematic decision pathways that prioritize visual clutter detection over content-based factors. The MLLM decision tree establishes badge clutter (Q7) as the primary discriminator, followed by color hierarchy (Q2) and price clarity (Q9), representing a structured approach to complexity assessment. This systematic evaluation contrasts significantly with human intuitive patterns, where the most frequently cited complexity reason ("products look similar") corresponds to lower-priority decision paths in the MLLM tree, indicating that diagnostic prompting enables more consistent visual assessment criteria.

\begin{table}[H]
\centering
\caption{Decision Tree Classification Rules for MLLM Diagnostic Assessment}
\label{tab:decision_rules}
\begin{tabular}{clc}
\toprule
\textbf{Path} & \textbf{Decision Rule} & \textbf{Predicted Class} \\
\midrule
1 & Q7 $\leq$ 0.5 $\land$ Q2 $\leq$ 0.5  & Complex \\
2 & Q7 $\leq$ 0.5 $\land$ Q2 > 0.5  & Not Complex \\
3 & Q7 > 0.5 $\land$ Q9 $\leq$ 0.5 $\land$ Q5 $\leq$ 0.5 & Complex \\
\bottomrule
\end{tabular}
\end{table}

Table~\ref{tab:human_alignment} provides empirical validation by demonstrating correspondence between human complexity concerns and diagnostic questions. Human frequency data shows "products look too similar" (Q4) as the most cited reason, followed by "text is small or too much to read" (Q5), and "colors/highlights are too loud" (Q2). The alignment confirms that our diagnostic framework captures specific visual elements humans perceive as complex, though some diagnostic questions address technical factors not explicitly mentioned by human evaluators.

\begin{table}[H]
\centering
\caption{Human Complexity Factors and Diagnostic Question Alignment}
\label{tab:human_alignment}
\begin{tabular}{llc}
\toprule
\textbf{Human Complexity Reason} & \textbf{Diagnostic Question(s)} & \textbf{Feature Importance} \\
\midrule
Products look too similar & Q4: ASIN faceouts similarity & 5.5\% \\
Text is small or too much to read & Q5: ASIN title/price readability & 10.7\% \\
Colors/highlights are too loud & Q2: Sponsored badges/highlights & 22.1\% \\
Product boxes are packed together & Q6: ASIN faceout spacing & 7.1\% \\
Too many badges, icons or labels & Q7: Badge clutter assessment & 38.6\% \\
Products seem irrelevant & Q15: Result relevance & <0.1\% \\
Filter section looks crowded & Q3: Navigation panel density & <0.1\% \\
\bottomrule
\end{tabular}
\end{table}

Cross-referencing Tables~\ref{tab:decision_rules} and \ref{tab:human_alignment} reveals a systematic divergence: humans prioritize content similarity (Q4), while MLLMs emphasize visual clutter factors (Q7: 38.6\%, Q2). The negligible importance of Q15 and Q3 (<0.1\%) suggests content-based assessments provide limited discriminative value compared to layout-focused questions. These patterns indicate that effective diagnostic frameworks should prioritize visual design questions (Q2, Q6, Q7) over content relevance assessments for MLLM-based evaluation.

\subsection{Qualitative Validation Through Failure Cases}

\begin{figure}[H]
% \label{figure:error_cases}
  \centering
  \begin{minipage}[t]{0.48\textwidth}
    \centering
    \includegraphics[width=\textwidth]{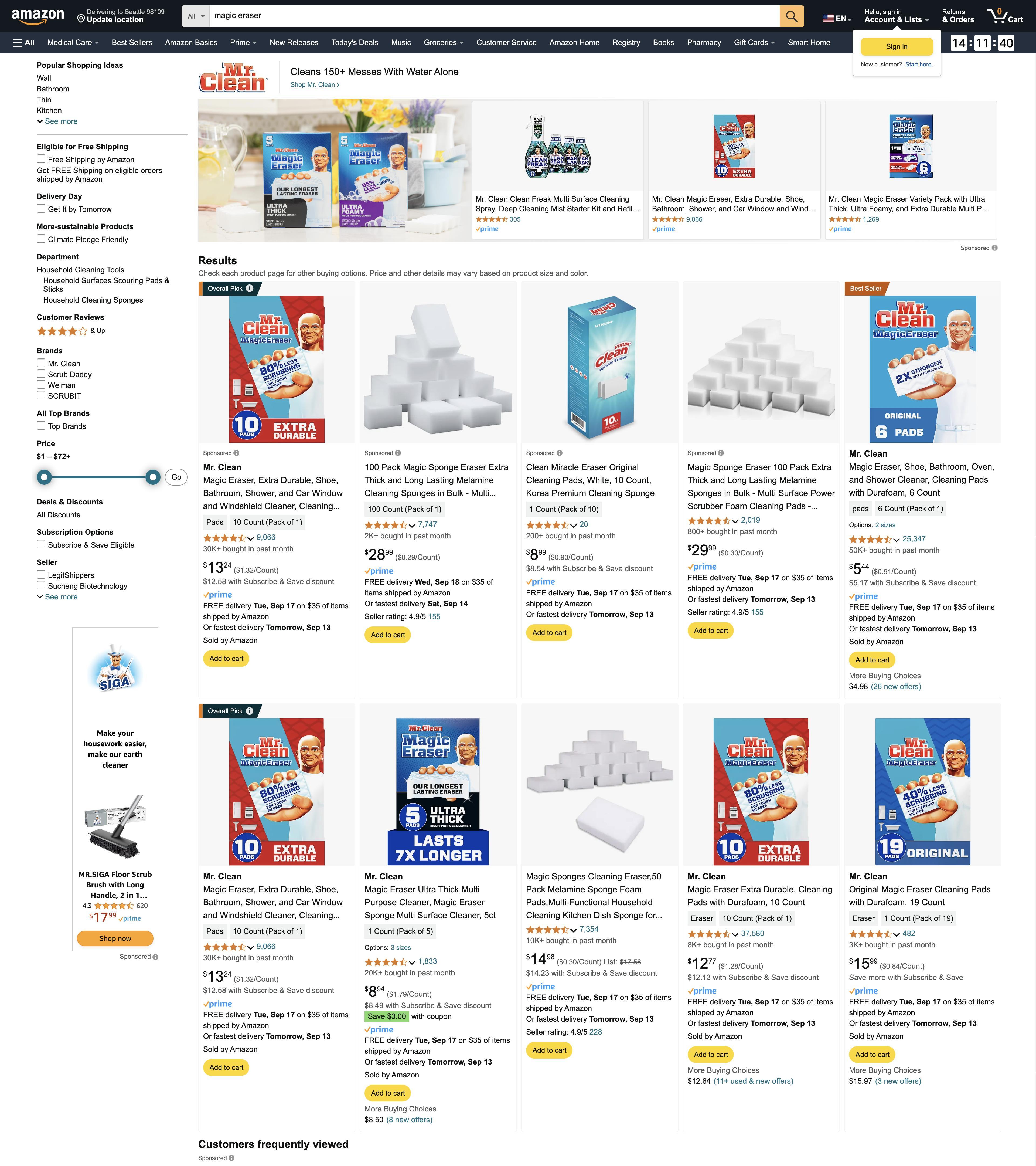}
    \caption*{\textbf{Case 1: "Magic Eraser"}}
    \small
    
    \textbf{Human:} Complex\\
    % \textbf{Reasons:} Products too similar, Text density \\
    \textbf{MLLM:} Not Complex\\
  \end{minipage}
  \hfill
  \begin{minipage}[t]{0.48\textwidth}
    \centering
    \includegraphics[width=\textwidth]{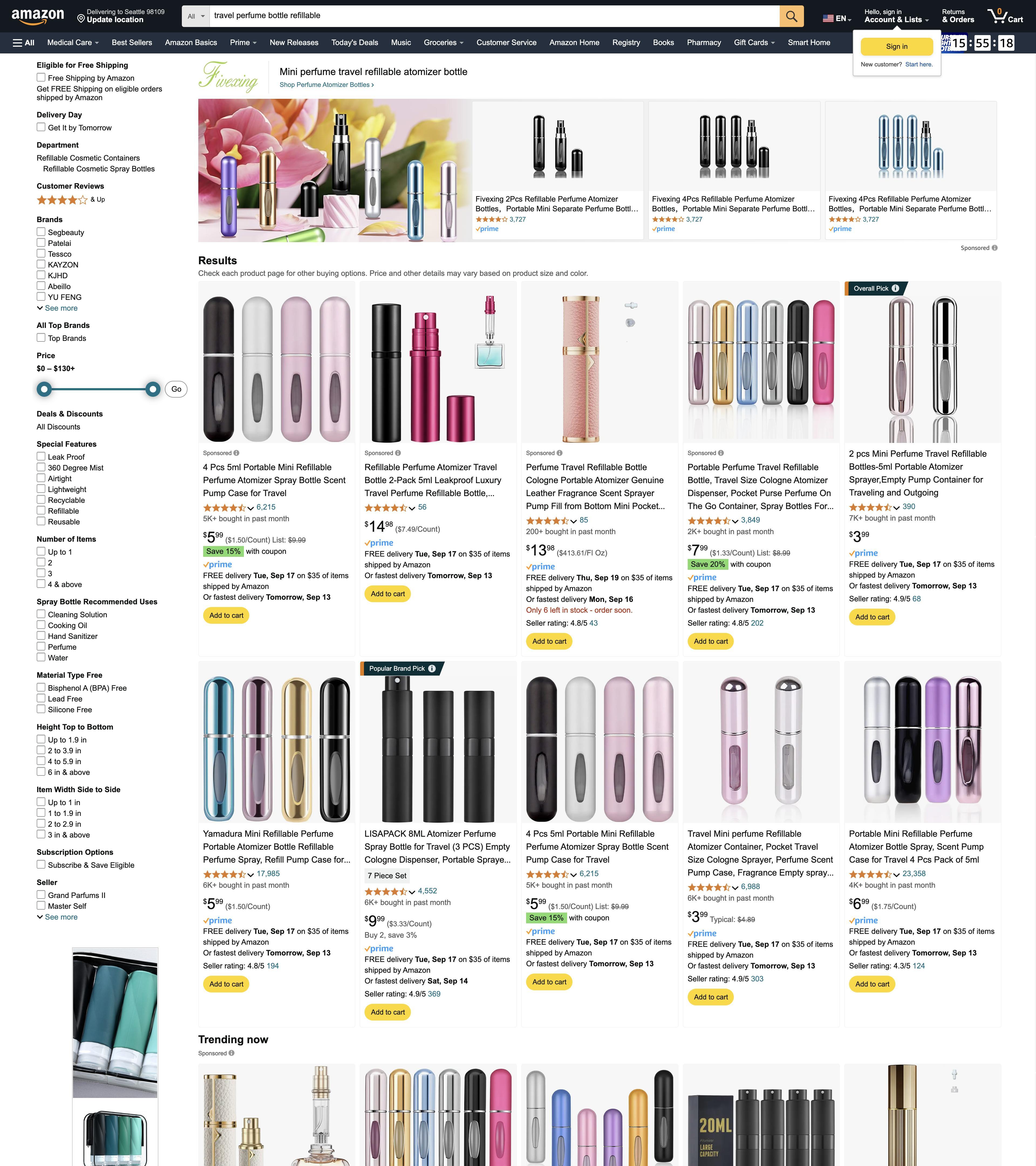}
    \caption*{\textbf{Case 2: "Travel Perfume Bottle"}}
    \small
    
    \textbf{Human:} Complex 
    % \textbf{Reasons:} Loud colors, Products similar, Text density\\
    
    \textbf{MLLM:} Not Complex\\
    
  \end{minipage}
  \caption{MLLM Evaluator Failure Cases}
  \label{fig:failure_cases}
\end{figure}

While diagnostic prompting shows numerical improvements over standard approaches, systematic evaluation limitations remain evident. Figure~\ref{fig:failure_cases} demonstrates persistent perceptual gaps that reveal current constraints in MLLM visual complexity assessment. In both cases, humans unanimously identified these SRPs as visually complex: Case 1 due to product similarity and text density, while Case 2 was driven by loud colors, product similarity, and text density. However, MLLMs failed to identify complexity factors and classified both cases as not visually complex.

\section{Discussion}

\subsection{Preliminary Findings and Framework Assessment}

Our diagnostic framework demonstrates meaningful relative improvements in MLLM performance metrics, with F1-score improving from 0.031 to 0.297. However, these findings require substantial future work before broader adoption. The framework's reliance on a single domain (Amazon SRP) and limited sample size (n=200) constrains generalizability. While the human-aligned question approach demonstrates promise, the Cohen's kappa of 0.071 remains low and far from acceptable moderate to substantial agreement levels (0.6-0.8).

\subsection{Critical Limitations and Alignment Gaps}

While these improvements are encouraging, the absolute performance levels and limited sample size indicate that substantial development remains before practical deployment. The decision tree analysis reveals fundamental differences in reasoning pathways for disagreement cases between human evaluators and MLLMs using our diagnostic approach, as demonstrated in the failure cases shown in Figure~\ref{fig:failure_cases}. While humans prioritize product similarity assessment, MLLMs emphasize badge clutter, suggesting partial rather than complete alignment. This discrepancy highlights a core challenge: our diagnostic questions capture some aspects of human reasoning but may miss critical perceptual patterns that influence complexity assessment. The framework's binary response structure, while improving consistency, potentially oversimplifies nuanced visual relationships that exist on continuous rather than discrete scales.

\subsection{Future Work}

To improve the low human-MLLM agreement, future work should test whether expanding from yes/no answers to 1-5 scale ratings for both human annotators and MLLMs helps better capture visual complexity nuances. Additionally, exploring fine-tunable MLLM models that can be trained on human ground truth complexity data could potentially improve alignment beyond what prompt-based approaches like Claude achieve.

The failure cases in Figure~\ref{fig:failure_cases} suggest two complementary prompting approaches. First, developing targeted questions that specifically address product similarity detection and color intensity assessment - for example, explicit comparison prompts like "Rate how visually similar these products appear" or color-focused questions like "Does the color scheme create visual noise." Alternatively, persona-based evaluation frameworks could address the inherent subjectivity in visual complexity assessment. Perception varies significantly across customer segments with different browsing behaviors, technical familiarity, and interface preferences.

Future work could develop persona-specific diagnostic questions that capture how different customer types perceive search page complexity. For example, mobile-first shoppers, power users, and accessibility-focused users may have different complexity thresholds. This would enable personalized MLLM evaluation aligned with target audience expectations rather than assuming universal complexity standards.

Expanding validation beyond Amazon SRPs to other e-commerce interfaces would establish whether our diagnostic questions generalize or require domain-specific adaptation. Testing with larger datasets (10K+ images) would also help determine if the current performance gains hold at scale and whether statistical significance can be achieved.

\section{Conclusion}

This work contributes to the growing field of MLLM evaluation by suggesting that structured diagnostic approaches may improve human-model alignment in subjective visual assessment tasks. Our findings advance understanding of how prompting methodologies influence MLLM reliability and provide insights into the fundamental differences between human and MLLM reasoning.

The systematic evaluation framework developed here offers a methodology for grounding MLLM evaluation in empirical human reasoning patterns. By translating human complexity assessment into structured diagnostic questions, we establish a pathway for improving model performance without requiring computationally intensive fine-tuning. This represents a practical contribution to prompt engineering research in multimodal contexts.

Our analysis reveals that while numerical performance improvements are achievable, fundamental perceptual gaps between humans and current MLLMs persist. The divergent prioritization of visual design elements versus content factors suggests that achieving true human-MLLM alignment requires deeper understanding of human visual cognition processes. This indicates that diagnostic prompting, while promising, represents an initial step rather than a complete solution.

For practitioners developing automated evaluation systems, this research demonstrates both the potential and current boundaries of MLLM-based assessment. The systematic approach developed here provides a foundation for future work addressing scalability and reliability challenges in subjective visual evaluation domains. Continued research with larger datasets and refined methodologies will be essential for advancing this work.

\section*{Acknowledgments}

We thank the volunteer annotators with search domain expertise who participated in the human annotation task, making this analysis possible. We also thank Tom Blake and Kailin Clarke for their valuable quality reviews and feedback that improved this work.

% Bibliography
\bibliographystyle{plain}
\bibliography{bibliography}

% Include appendix
\appendix

\clearpage

\appendix

\section*{Appendix}
\section{Standard Gestalt-Based Prompt}
\label{app:standard_prompt}

This section presents the complete traditional Gestalt-based prompt used as the baseline for single-shot MLLM evaluation. This approach relies on theoretical principles without human frequency validation. The prompt below shows exactly what is fed to the MLLM as a single input.

\begin{lstlisting}[
    breaklines=true,
    breakatwhitespace=true,
    basicstyle=\tiny\ttfamily,
    frame=single,
    frameround=tttt,
    linewidth=\textwidth,
    postbreak=\mbox{\textcolor{red}{$\hookrightarrow$}\space},
    showstringspaces=false,
    columns=flexible,
    caption={Complete Standard Gestalt-Based Prompt}
]


### Instructions: 
You are an HCI researcher and expert in visual psychology, design, and user experience. You have extensive knowledge of the Gestalt principles of perception and their application in visual design and user experience.

In your role as an HCI expert, you will take one or more (maximum 3) images of screenshots of a search page. All the screenshots would belong to the same search page scrolled to the bottom to include all the results. Hence you have to imagine them being screenshot of the same search webpage cropped at different heights. To get a view of the entire search page, imagine that they are stitched one below the other (in the same chronological order as they are provided in the input). The first image will be placed at the top and the second image (if provided) will be combined below the first image, the third image (if provided) will be combined below the second and so on. This is how a person would see these images on a webpage.

Your task is to analyze the visual complexity of this page layout based on the following Gestalt principles of perception:

1. Law of Similarity: Objects that share similar characteristics, such as color, shape, size, or texture, are perceived as belonging to a group or pattern.

2. Law of Proximity: Objects that are close together are perceived as being part of a group or pattern.

3. Law of Pragnanz (Good Figure/Law of Simplicity): The mind tends to perceive objects in their simplest, most stable form, which makes the object appear as a complete, unified whole.

4. Law of Closure: The mind has a tendency to perceive incomplete figures as complete, whole objects by filling in the missing information.

5. Law of Continuity: The mind perceives elements that are arranged in a continuous pattern, line, or curve as belonging together and being a single form.

6. Law of Figure/Ground: The mind separates objects into two distinct categories: the figure (the object of focus) and the ground (the background or surrounding area).

After analysing using the Gestalt principles, you will score the whole page based for each principle based on the scoring system below.
The scoring should be based on the overall page layout, considering the collective arrangement and presentation of all product images and surrounding elements.

Scoring: 

Law of Similarity:
1 point - All products are identical in color, shape, size, texture, and every other visual aspect.
2 points - Products vary in only one aspect (e.g., color, shape, size, or texture), but are otherwise identical.
3 points - Products vary in two aspects (e.g., color and shape, size and texture, etc.).
4 points - Products vary in three aspects (e.g., color, shape, and size).
5 points - Each product is completely unique, differing in color, shape, size, texture, and all other visual aspects.

Law of Proximity: 
1 point - Similar items are placed immediately adjacent to each other, with no visible spacing in between. This close proximity makes it extremely challenging to differentiate the items.
2 points - Similar items have some spacing in between, but the spacing is minimal, causing the items to be perceived as part of a group or cluster. Differentiation requires conscious effort.
3 points - Similar items have sufficient spacing in between, allowing each item to be clearly distinguished as a separate, individual entity without any perceived grouping.

Law of Pragnanz (Good Figure/Law of Simplicity):
1 point - The page layout appears highly complex overall due to a combination of high similarity across products and intricate individual designs with many distinct elements/shapes, making it extremely challenging to perceive clear, distinct figures.
2 points - The page layout has a relatively high level of complexity due to moderate similarity across products and moderately complex individual designs (4-5 distinct elements/shapes), hindering the ability to easily discern the simplest form or figure for each item.
3 points - The page layout has a moderate level of complexity, with a mix of similar and diverse products, and a range of simple to moderately complex individual designs, allowing for some perception of distinct figures, but with moderate difficulty.
4 points - The page layout appears relatively simple overall due to a good level of diversity across products and predominantly simple individual designs (2-3 distinct elements/shapes), making it easier to perceive each item as a clear, distinct figure with discernible form and details.
5 points - The page layout appears highly simple overall, with a high level of diversity across products and all individual designs being single, unified shapes or forms, enabling viewers to quickly and easily discern the simple, unified figure of each unique item.

Law of Closure:
1 point - The arrangement of products appears completely random or chaotic, with no discernible pattern or alignment, significantly disrupting the visual flow and continuity.
2 points - The arrangement exhibits some alignment or structure, but with frequent disruptions or irregularities that break the continuity and make it difficult for the eye to follow a consistent path.
3 points - The products are arranged in a mostly consistent pattern or alignment, with minor disruptions or breaks, allowing for a moderate degree of visual continuity and the ability to perceive some continuous paths.
4 points - The arrangement of products forms clear, continuous patterns or alignments, with minimal disruptions, enabling the eye to easily follow distinct visual paths and maintain a strong sense of continuity.
5 points - The products are arranged in a perfect, unbroken pattern or alignment, providing seamless visual continuity without any disruptions, allowing the eye to effortlessly follow a single, continuous path throughout the entire layout.

Law of Continuity:
1 point - The arrangement of elements appears completely random or chaotic, with no discernible pattern, line, or curve. The viewer's eye cannot follow any continuous path or flow, resulting in a highly disjointed and fragmented visual experience.
2 points - While there may be some semblance of a continuous pattern or arrangement, it is weak and frequently interrupted by significant breaks or disruptions. The continuity is consistently undermined, making it challenging for the viewer's eye to transition smoothly between elements.
3 points - The elements are arranged in a recognizable pattern or alignment, but with several noticeable interruptions or discontinuities that disrupt the visual flow at various points. The viewer's eye can follow the general continuity, but the path is frequently broken or fragmented.
4 points - The arrangement of elements exhibits a clear and dominant continuous pattern, line, or curve, with only minor, negligible disruptions that do not significantly impede the overall sense of continuity. The viewer's eye can transition smoothly between elements while following the distinct visual path.
5 points - The elements are arranged in a flawless, unbroken pattern, line, or curve, creating an extremely smooth and seamless visual flow without any interruptions or breaks. The viewer's eye can effortlessly follow the continuous path from one element to the next, resulting in a highly coherent and unified visual experience.

Law of Figure/Ground:
1 point - The layout lacks clear figure-ground relationships, with elements blending into the background or competing for attention, making it extremely difficult to discern distinct figures against a coherent ground.
2 points - The figure-ground relationships are weak, with elements having a low level of differentiation from the background, leading to significant visual tension and ambiguity in identifying distinct figures.
3 points - There is a moderate level of figure-ground distinction, with some elements standing out as figures against the background, but others blending in or competing for attention, resulting in a somewhat inconsistent visual experience.
4 points - Most elements establish clear figure-ground relationships, with distinct figures standing out prominently against a coherent background, facilitating easy visual differentiation and recognition of individual items.
5 points - All elements exhibit exceptional figure-ground relationships, with each item clearly defined as a distinct figure that stands out prominently and unambiguously against a unified background, providing optimal visual differentiation and ease of recognition.

Overall assessment. Based on the points and comments along the above directives provide final assessment for the input image.
Result: Final visual complexity score ranging from 1 to 5 points based on the above assessment and conclusions. 

First, I provide you with an example screenshot of search page of cactus dancing toys, that are related to a query "dancing cactus toy", and then i provide sample output explanations with scores for each principle. Based on this example I will then provide input images for which score you will have to explain and score the screenshots of search pages using the Gestalt Principles to assess visual complexity in them.

### Sample Input:
\end{lstlisting}

\section{Human-Aligned Diagnostic Prompt}
\label{app:diagnostic_prompt}

This section presents our complete human-aligned diagnostic prompting framework, developed through systematic analysis of human complexity reasoning patterns. The prompt below shows exactly what is fed to the MLLM as a single input, demonstrating the structured two-part evaluation process.
\begin{lstlisting}[
    breaklines=true,
    breakatwhitespace=true,
    basicstyle=\tiny\ttfamily,
    frame=single,
    frameround=tttt,
    linewidth=\textwidth,
    postbreak=\mbox{\textcolor{red}{$\hookrightarrow$}\space},
    showstringspaces=false,
    columns=flexible,
    caption={Complete Human-Aligned Diagnostic Prompt}
]
### Instructions: 
You are a visual layout analysis assistant evaluating a screenshot of an Amazon search result page. Your task has two parts:

1. Perform a neutral diagnostic evaluation of the layout using 25 visual UX questions.
2. Then assign visual complexity scores and explanations.

Use only the screenshot to answer layout-related questions. Do not infer category-specific logic or product content.

## BUSINESS CONTEXT & TERMINOLOGY

Before beginning the evaluation, understand these Amazon-specific terms:

**SRP (Search Results Page)**: The main page displaying search results with products, ads, and navigation elements.

**ATF (Above-the-Fold)**: The visible area of the page before scrolling, typically containing sponsored brand widgets and top banner advertising.

**ASIN Faceouts**: Individual product listing tiles/cards showing product image, title, price, ratings, and other metadata. These are the core product presentation units.

**Sponsored Products**: Paid advertising placements that appear within organic search results, marked with "Sponsored" labels.

**Sponsored Brand Widget**: Large banner advertisements typically appearing at the top of search results (ATF area).

**RIB (Refinements Info Bar / Result Info Bar)**: The horizontal bar that appears above search results, providing quick access to product type pills and popular filters. It helps users narrow results and provides an overview of product categories for broad queries.

**Navigation Panel**: Left sidebar containing filters, refinements, categories, and search narrowing options.

**Prime Callouts**: Visual indicators showing Prime eligibility (Prime logo, shipping benefits).

**Deal Highlights**: Visual elements showing discounts, savings, lightning deals, and promotional pricing.

## PART 1: DIAGNOSTIC EVALUATION

You are acting as a neutral UX auditor evaluating Amazon SRP layout structure.

**Instructions:**
- Answer each of the 25 questions below using Yes / No / Not Sure
- Focus only on visual layout structure, flow, spacing, repetition, and clarity
- Be objective - do not apply customer preferences or product-specific logic

### Above-the-Fold & Header Section

**Q1.** Does the ATF Sponsored Brand widget or top banner advertising feel too prominent or oversized compared to organic results?

**Q2.** Are the Sponsored Products badges, deal highlights, Prime callouts, or color schemes too loud, bright, or distracting?

### Left Navigation Panel

**Q3.** Does the left navigation panel with filters, refinements, and categories look crowded or overwhelming?

### ASIN Faceouts - Product Grid

**Q4.** Do the ASIN faceouts in the search results look too similar to each other in terms of images, styling, or presentation?

**Q5.** Is the ASIN title, price, review text, or product details too small, dense, or difficult to read comfortably?

**Q6.** Do the ASIN faceouts appear packed too closely together without adequate white space or breathing room?

**Q7.** Are there too many badges (Best Seller, Amazon's Choice, Deal, Prime) cluttering the ASIN faceouts?

**Q8.** Are ASIN hero images inconsistent in size, background style, or photographic quality across the result set?

**Q9.** Do the price formats, strikethrough pricing, savings callouts, and deal information create visual confusion?

**Q10.** Is there poor alignment between ASIN images, titles, prices, and ratings within individual faceouts?

**Q11.** Are there inconsistent fonts, text sizes, or typography styles across different ASIN faceouts?

**Q12.** Do the ASIN thumbnail images vary too much in style, cropping, or presentation across the result set?

### Sponsored Products Integration

**Q13.** Do Sponsored Products blend too much with organic search results, making ad content unclear?

### Search Results Organization

**Q14.** Do the organic search results appear unorganized, randomly ordered, or lack clear sorting logic?

**Q15.** Do the returned ASINs seem irrelevant, unrelated, or poorly matched to the search query?

### RIB & Page Transitions

**Q16.** Are there jarring visual transitions between search results and recommendation widgets or other sections?

**Q17.** Do different page sections (navigation panel, results, RIB) lack clear visual separation or boundaries?

### Overall SRP Layout & Hierarchy

**Q18.** Is the visual hierarchy unclear for natural F-pattern reading flow on the SRP?

**Q19.** Does the overall SRP feel overwhelming due to high information density in ASIN faceouts?

**Q20.** Is the overall ASIN faceout grid layout visually unbalanced or lopsided in weight distribution?

### Interactive Elements & Affordances

**Q21.** Are there too many competing calls-to-action, buttons, or interactive elements fighting for attention?

**Q22.** Are clickable elements (Add to Cart, wishlist, compare) visually indistinguishable from static content?

**Q23.** Do skeleton screens, lazy-loaded ASIN images, or missing content create visual disruption or incompleteness?

**Q24.** Does the SRP layout violate expected Amazon e-commerce patterns or user interface conventions?

### General Complexity Assessment

**Q25.** Are there other visual complexity factors not covered by specific questions that make the SRP feel overwhelming?

## PART 2: VISUAL COMPLEXITY EVALUATION

Use your diagnostic answers from Part 1 to assign complexity scores for each persona:

**Complexity Score Scale:**
- 1 = Extremely complex and cluttered
- 2 = Very difficult to navigate visually  
- 3 = Somewhat complex but manageable
- 4 = Mostly clean and clear
- 5 = Very easy to scan, visually simple

**IMPORTANT:** Return only the JSON output. Do not include any explanation or formatting outside the JSON block. Ensure proper JSON formatting - no trailing commas, no missing commas.

{
  "diagnostics": {
    "Q1": "Yes",
    "Q2": "No",
    ...
    "Q25": "Yes"
  },
  "complexity_score": 2,
  "explanation": "The page feels dense and overwhelming with too many competing visual elements making it difficult to scan quickly."
}
\end{lstlisting}

\section{Rationale for Human-Aligned Diagnostic Prompting}
\label{app:rationale_for_diagnostic_prompt}

The central motivation for diagnostic questioning stems from a fundamental human-MLLM alignment problem in visual complexity assessment. Humans evaluate complexity through intuitive, experience-based pattern recognition that integrates multiple contextual factors simultaneously, while MLLMs rely on structured input and lack implicit contextual understanding. Traditional Gestalt-based approaches fail because they assume theoretical principles directly translate to human complexity perception, achieving F1-scores of only 0.031 with Cohen's Kappa of -0.016 (worse than random chance). This theoretical versus empirical gap reveals itself in how Gestalt principles do not align with how humans actually reason about complexity in e-commerce contexts, while six broad principle scores provide insufficient granularity to identify specific complexity factors driving human perception.

Our human-aligned diagnostic framework addresses these limitations through three key innovations that work synergistically to improve MLLM performance. The empirical foundation ensures questions derive from systematic analysis of actual human complexity reasoning patterns rather than theoretical assumptions. SRP-specific terminology uses Amazon vocabulary including ASIN faceouts, Sponsored Products, and Prime callouts for improved MLLM contextual understanding that bridges the domain knowledge gap. Granular assessment through 25 specific diagnostic questions enables detailed complexity factor identification, achieving F1-scores of 0.297 (+858\% improvement) and Cohen's Kappa of 0.071 (positive human-MLLM alignment). This approach maintains theoretical rigor through Gestalt principle alignment while ensuring practical effectiveness and represents a novel methodological pathway for improving MLLM performance in subjective assessment tasks across domains.

\section{Question Development Methodology}
\label{app:question_development}

Our systematic process began with comprehensive human complexity reason collection from 4-7 volunteer annotators with search domain expertise per query across 200 Amazon SRP samples, followed by broad complexity factor identification of general concerns such as products appearing too similar, colors being too loud, or text being hard to read. The core innovation involved SRP-visual elaboration, translating each broad human concern into specific diagnostic questions tied to concrete visual elements that MLLMs can observe and evaluate. For example, "products too similar" became Q4 asking "Do ASIN faceouts look too similar in images, styling, or presentation?" while "text hard to read" transformed into Q5 inquiring "Is ASIN title, price, review text too small, dense, or difficult to read?" and "colors too loud" evolved into Q2 questioning "Are Sponsored Products badges, deal highlights, Prime callouts too loud or distracting?"

To improve MLLM contextual understanding, we incorporated authentic Amazon SRP vocabulary including ASIN Faceouts, Sponsored Products, Prime Callouts, ATF Sponsored Brand Widget, Navigation Panel, and Deal Highlights rather than generic e-commerce terminology. Each diagnostic question maintains theoretical grounding through alignment with established Gestalt principles, with Q4 reflecting the Similarity Principle, Q5 embodying Figure-Ground Relationships, Q6 corresponding to the Proximity Principle, and Q7 connecting to Simplicity/Prägnanz. Our methodology was validated through decision tree feature importance analysis using 5-fold stratified cross-validation, revealing that Q7 (badge clutter) achieved 38.6\% importance as the highest predictor despite ranking only 8th in human frequency, Q2 (text readability) maintained 10.7\% importance, and Q13 (price display) demonstrated 8.5\% importance as strong predictors of complexity perception.

\section{Key Differences: Standard vs. Human-Aligned Approaches}
\label{app:comparative_analysis}

This comparative analysis highlights the fundamental distinctions between traditional Gestalt-based evaluation and our human-aligned diagnostic framework.

\subsection{Methodological Differences}

\begin{table}[H]
\centering
\begin{tabular}{|p{3cm}|p{5cm}|p{5cm}|}
\hline
\textbf{Aspect} & \textbf{Standard Gestalt} & \textbf{Human-Aligned Diagnostic} \\
\hline
\textbf{Foundation} & Theoretical Gestalt principles without empirical validation & Systematic analysis of actual human complexity reasoning patterns \\
\hline
\textbf{Terminology} & Generic visual design vocabulary & Amazon-specific SRP terminology (ASIN faceouts, Sponsored Products, etc.) \\
\hline
\textbf{Assessment Granularity} & 6 broad principle scores (Similarity, Proximity, etc.) & 25 specific diagnostic questions targeting granular complexity factors \\
\hline
\textbf{Question Weighting} & All principles treated equally & Frequency-weighted prioritization based on human citation data \\
\hline
\textbf{Evaluation Approach} & Single-shot aggregate scoring & Systematic diagnostic evaluation with binary responses \\
\hline
\end{tabular}
\caption{Methodological comparison between standard and human-aligned approaches}
\end{table}

\subsection{Performance Comparison}

\begin{table}[H]
\centering
\begin{tabular}{|p{3cm}|p{3cm}|p{3cm}|p{4cm}|}
\hline
\textbf{Metric} & \textbf{Standard Gestalt} & \textbf{Human-Aligned} & \textbf{Improvement} \\
\hline
\textbf{F1-Score} & 0.031 & 0.297 & +858\% improvement \\
\hline
\textbf{Precision} & 0.060 & 0.110 & +83\% improvement \\
\hline
\textbf{Recall} & 0.021 & 0.309 & +1,371\% improvement \\
\hline
\textbf{Cohen's Kappa} & -0.016 & 0.071 & From worse-than-random to positive agreement \\
\hline
\textbf{Statistical Significance} & Not achieved & p = 0.2912 (McNemar's test) & Not statistically significant \\
\hline
\textbf{Human Alignment} & Worse than chance & Moderate agreement & Substantial improvement \\
\hline
\end{tabular}
\caption{Performance metrics comparison between approaches}
\end{table}

\subsection{Theoretical vs. Empirical Grounding}

\begin{table}[H]
\centering
\begin{tabular}{|p{3cm}|p{5cm}|p{5cm}|}
\hline
\textbf{Grounding Aspect} & \textbf{Standard Gestalt} & \textbf{Human-Aligned Diagnostic} \\
\hline
\textbf{Foundation} & Assumes theoretical principles directly translate to complexity perception & Grounds evaluation in empirically observed human reasoning patterns \\
\hline
\textbf{Validation Method} & Lacks validation against actual human patterns & Validates through frequency analysis and predictive modeling \\
\hline
\textbf{Scoring System} & Abstract scoring may not align with real-world factors & Specific, actionable complexity factors for improvement \\
\hline
\textbf{Actionability} & Limited actionable insights for interface optimization & Maintains theoretical rigor while ensuring practical relevance \\
\hline
\textbf{Predictive Power} & Poor correlation with human judgment & Strong predictive correlation with human complexity assessment \\
\hline
\textbf{Research Rigor} & Theoretical assumptions without empirical testing & Systematic empirical validation with statistical significance \\
\hline
\end{tabular}
\caption{Theoretical versus empirical grounding comparison}
\end{table}

\subsection{Practical Implementation Differences}

\begin{table}[H]
\centering
\begin{tabular}{|p{3cm}|p{5cm}|p{5cm}|}
\hline
\textbf{Implementation Aspect} & \textbf{Standard Gestalt} & \textbf{Human-Aligned Diagnostic} \\
\hline
\textbf{Prompt Structure} & Single prompt with broad instructions & Structured two-part evaluation process \\
\hline
\textbf{MLLM Requirements} & Requires interpretation of abstract principles & Provides specific, contextual diagnostic questions \\
\hline
\textbf{Output Format} & Produces aggregate scores with limited value & Generates detailed complexity factor identification \\
\hline
\textbf{Diagnostic Value} & Difficult to identify specific improvement areas & Enables targeted interface optimization recommendations \\
\hline
\textbf{Evaluation Complexity} & Simple but ineffective single-pass evaluation & Systematic diagnostic with actionable insights \\
\hline
\textbf{Domain Adaptation} & Generic approach requires manual adaptation & SRP-specific terminology built-in \\
\hline
\end{tabular}
\caption{Practical implementation differences between approaches}
\end{table}

\subsection{Scalability and Reliability}

\begin{table}[H]
\centering
\begin{tabular}{|p{3cm}|p{5cm}|p{5cm}|}
\hline
\textbf{Scalability Factor} & \textbf{Standard Gestalt} & \textbf{Human-Aligned Diagnostic} \\
\hline
\textbf{Consistency} & High variability in evaluation across instances & Structured diagnostic questions reduce evaluation variability \\
\hline
\textbf{Interpretability} & Abstract scores difficult to interpret & Binary responses and specific factors enable clear analysis \\
\hline
\textbf{Actionability} & Limited guidance for improvement & Granular diagnostic results support targeted improvements \\
\hline
\textbf{Validation} & No mechanism for ongoing validation & Empirical grounding ensures continued relevance to human perception \\
\hline
\textbf{Large-scale Deployment} & Inconsistent results at scale & Reliable performance across diverse SRP layouts \\
\hline
\textbf{Quality Assurance} & Difficult to validate output quality & Clear diagnostic framework enables quality verification \\
\hline
\end{tabular}
\caption{Scalability and reliability comparison for large-scale assessment}
\end{table}

\end{document}